\documentclass[]{spie}  
\usepackage{color}
\usepackage[]{graphicx}
\usepackage{amsfonts}

\title{Computational identification of significant actors \\ in paintings through symbols and attributes}

\author{David G.~Stork,\supit{a} Anthony Bourached,\supit{b} George H.~Cann,\supit{c} and Ryan-Rhys Griffiths\supit{d} 
\skiplinehalf 
\supit{a}Portola Valley, CA 94028 USA  \\
\supit{b}Oxia Palus  \\
\supit{c} Department of Space and Climate Physics, University College London, London, UK \\
\supit{d}Department of Physics, Cambridge University, Cambridge, UK}

\authorinfo{Send correspondence to the first author at {\tt artanalyst@gmail.com}.}

\begin{document} 
\maketitle 

\begin{abstract}
The automatic analysis of fine art paintings presents a number of novel technical challenges to artificial intelligence, computer vision, machine learning, and knowledge representation quite distinct from those arising in the analysis of traditional photographs.  The most important difference is that many realist paintings depict stories or episodes in order to convey a lesson, moral, or {\em meaning}.  One early step in automatic interpretation and extraction of meaning in artworks is the identifications of figures (\lq\lq actors\rq\rq ).  In Christian art, specifically, one must identify the actors in order to identify the Biblical episode or story depicted, an important step in \lq\lq understanding\rq\rq\ the artwork.  We designed an automatic system based on deep convolutional neural networks and simple knowledge database to identify saints throughout six centuries of Christian art based in large part upon saints\rq\ symbols or {\em attributes}.  Our work represents initial steps in the broad task of automatic semantic interpretation of messages and meaning in fine art.
\end{abstract}


\section{INTRODUCTION AND BACKGROUND} \label{sec:Intro}

The problem of even basic automatic interpretation of fine art paintings and drawings is extremely challenging and is quite different from  semantic image analysis of traditional photographs.  In most current research on automatic segmentation, object recognition, image captioning, question answering, and analysis of photographs, the output describes the state of affairs in a physical scene.  Many paintings, drawings and murals, particularly in the Western canon, were created to tell stories, convey lessons, or impute morals that have no counterpart in the vast research on analysis of natural photographs.    Much narrative art---and specifically most Western religious art---depicts episodes or stories that have a crafted or selected {\em meaning}.  Such meanings are expressed in natural language, as for instance \lq\lq Christ died for your sins,\rq\rq\ \lq\lq You should be willing to sacrifice your own son if your god demands it,\rq\rq\ \lq\lq Extend charity to those less fortunate,\rq\rq\ \lq\lq The love of money is the root of all evil,\rq\rq\ and innumerable others.  Full semantic analysis of such artwork must eventually extract or infer such meanings.

Artists often employ conventions, deliberate physical inconsistencies, highly stylized renderings, and more in order to help convey such messages and meanings.  Most artists, especially in the last millennium, use content, composition, medium, color, style, visual metaphors, and conventions such as symbols, \lq\lq invisible\rq\rq\ entities, and other techniques in the service of these goals.  Many of these properties of art have no counterpart in traditional natural photographs or videos and for this reason the techniques for the analysis of common photographs are of only modest value in automatic approaches to interpreting fine art.  
 
The understanding of meaning in much representational Western art, such as religious art, builds upon the identification of depicted figures or \lq\lq actors.\rq\rq\ \ For instance, the meaning in Vincenzo Catena\rq s {\em Christ giving the keys to Saint Peter} is lost on viewers who cannot recognize the central figures, that the keys symbolize the requirements for entry into heaven, and that the three women in the background are allegories of Faith, Hope, and Charity.  Traditional automatic facial recognition is nearly useless in this task, as there is no \lq\lq ground truth\rq\rq\ exemplar images of important figures.  Moreover, the stylistic variations among painters are far too great for current image recognition systems trained with natural photographs.  No current face recognition system is accurate for faces as diverse as those that appear in the works of Duccio, Leonardo, El Greco, Karel Apel, Jean Dubuffet, Willem de Kooning, among many others.  Art analysis differs methodologically from that for traditional photographs as well in that the total number of artworks available for training automatic systems, while large, is but a small fraction of the number of photographs available online and in curated databases that are vital to the success of state-of-the-art performance based on training deep neural networks.\cite{GoodfellowBengioCourville:16}  This disparity in corpora sizes is growing every day.

Central to the understanding of religious art such as Christian art in the Western canon are {\em symbols}, for instance a cross, lamb, keys, crown of thorns, ox and ass, and so on.  Furthermore, the non-physical conditions such as the upward floating of Christ in Rogier van der Weyden\rq s {\em Isenheim altarpiece}, the glowing radiant Christ child in Gerard David\rq s {\em Birth of Christ}, the halos appear in innumerable artworks, all conveying important and well-understood meanings.  Conventions such as the appearance of crepuscular rays (\lq\lq Rays of Buddha\rq\rq ) and golden heavenly rays have associations and thus convey meaning.  For all its aesthetic value, Rembrandt\rq s {\em Supper at Emmaus} would appear as a common dinner in a simple inn were it not for the radiant glow of the central figure, indicating his identity as Christ, and thus marking the scene as one of the central stories in the Bible.  One simply cannot understand such artworks and the artists\rq\ intentions without recognizing and reasoning about these conventions and their associations.\cite{deRynck:04,Zuffi:10}  Any automatic system for the interpretation of art must recognize and then make inferences using these unique artistic aspects.

One of the requisite early steps in interpreting such paintings, and thus automatic interpretation, is identifying the figures or \lq\lq actors,\rq\rq\ such as saints in Bible stories.  This is the task we address:  the automatic identification of major figures or \lq\lq actors,\rq\rq\ such as saints, in religious two-dimensional art, specifically paintings.  Major figures such as saints in Western religious art are frequently marked by the presence of a sign or {\em attribute}.\cite{Ferguson:90}  The general categories of {\em signs}---so-called {\em signifiers}---and their corresponding objects or concepts---so-called {\em signifieds}---is the central concern of the branch of philosophy known as {\em semiotics}.\cite{Aiello:19,Moriarty:05}  Thus Christ\rq s main attributes are the cross and the lamb ({\em Agnus Dei}), Saint Peter\rq s are keys, Saint Mark\rq s is the winged lion, God\rq s is a dove, and so on.  In religious iconography, an {\em attribute} is an object or symbol associated with a figure such as a saint, most often deriving from some episode or story from the Bible.  For instance, Saint Peter\rq s attribute, {\em keys}, stems from a passage in {\em Matthew} 16:19: 

\begin{quote}\lq\lq I will give to thee the keys of the kingdom of heaven. And whatsoever thou shalt bind upon earth, it shall be bound also in heaven: and whatsoever thou shalt loose upon earth, it shall be loosed also in heaven.\rq\rq\
\end{quote}

\noindent Some religious actors have several attributes (Saint Peter:  keys, boat, and fish), and several actors share the same attribute (Christ and John the Baptist:  cross), though which of multiple attributes is represented depends upon the settings, context, or episode being depicted.  Such symbols and attributes were widely used in religious paintings, altarpieces, stained-glass windows, prayer books, and so on, to help teach and reinforce Biblical stories, particularly to illiterate parishioners.\cite{Ferguson:90}  

\begin{figure}[h]
\begin{center}
\begin{tabular}{ccccc}
\includegraphics[height=.21\textwidth]{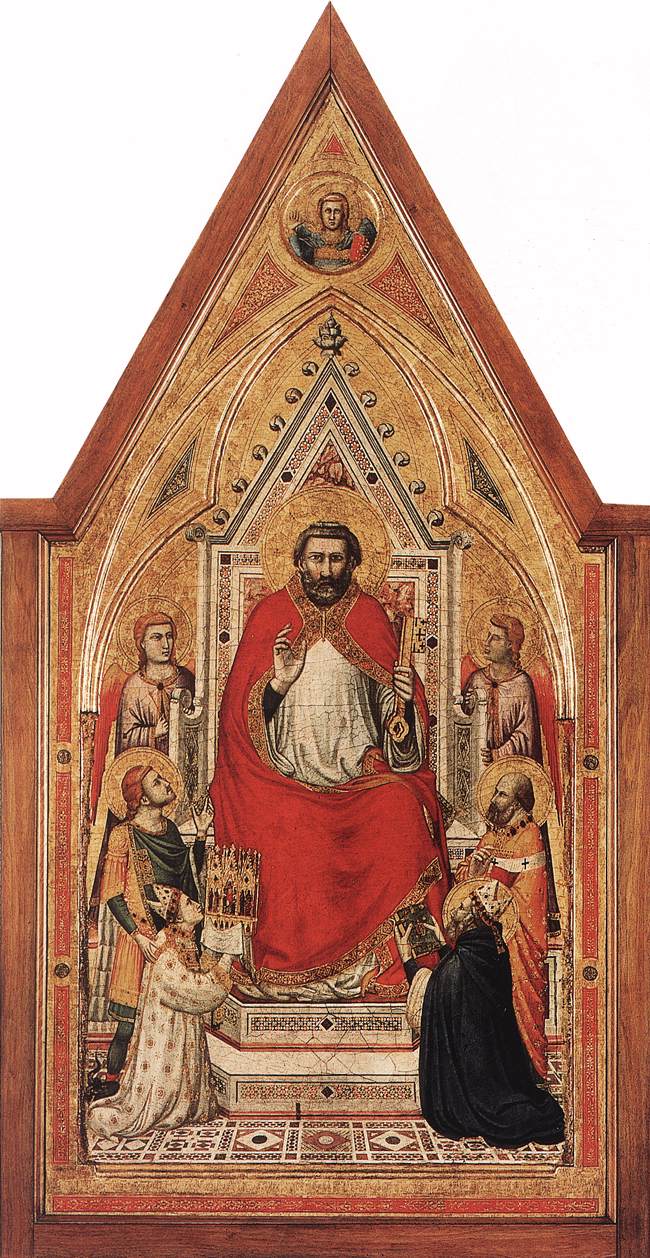} &
\includegraphics[height=.21\textwidth]{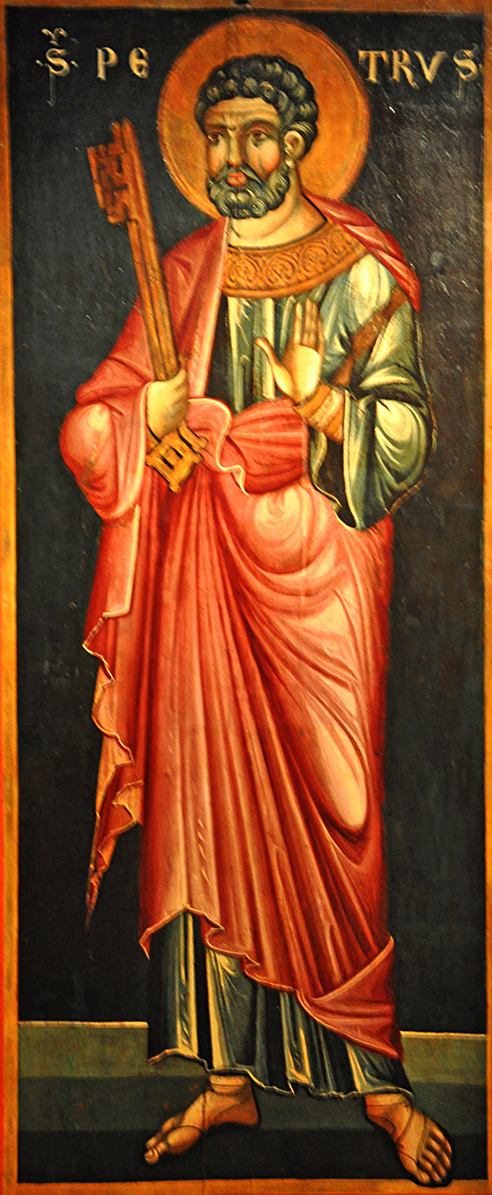} &
\includegraphics[height=.21\textwidth]{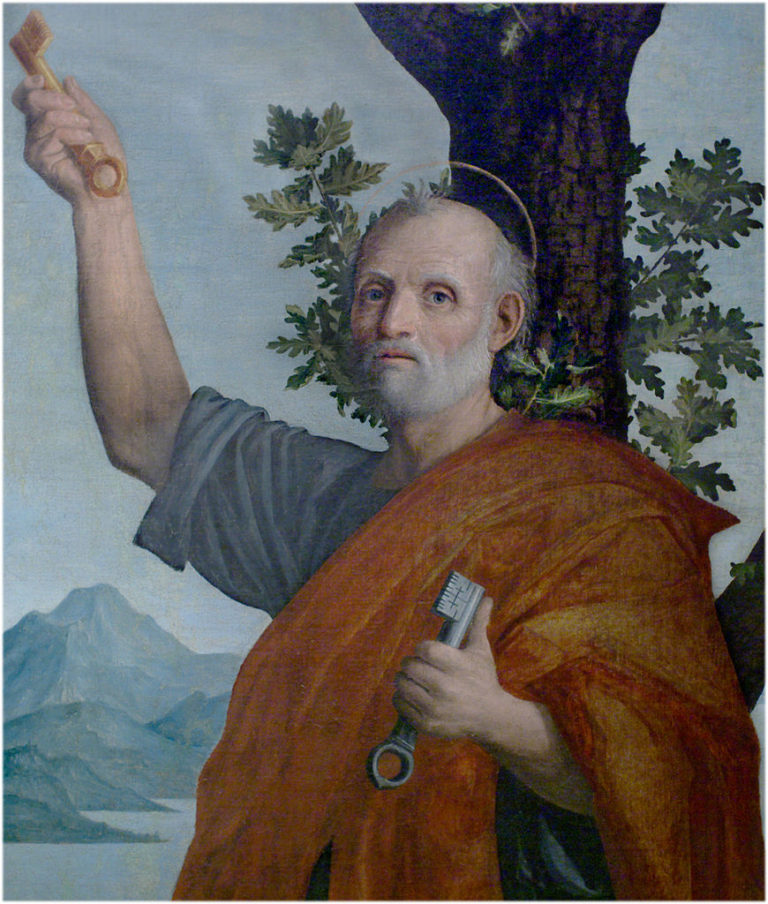} &
\includegraphics[height=.21\textwidth]{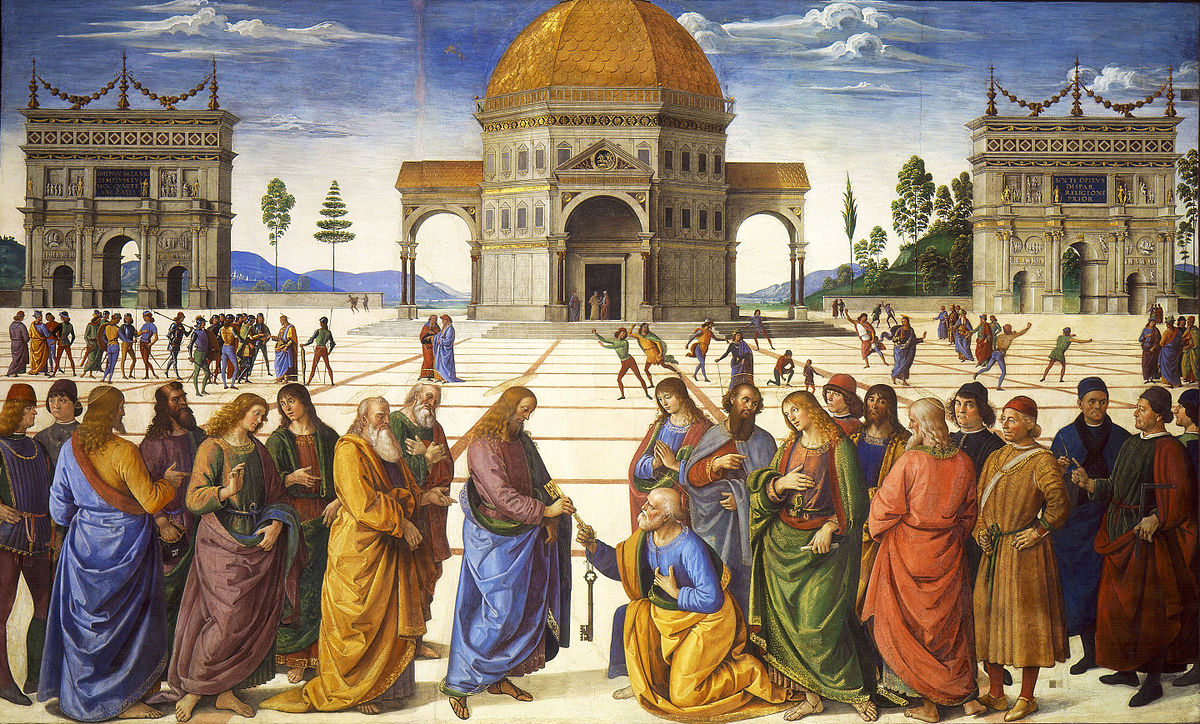} &
\includegraphics[height=.21\textwidth]{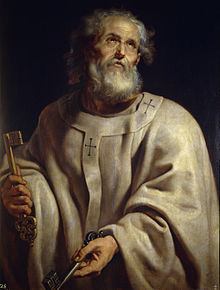} \\
a) & b) & c) & d) & e)
\end{tabular}
\end{center}
\caption{\label{fig:StPeter}a) Unknown artist\rq s {\em Stefani Triptych, Saint Peter enthroned} (c.~1330), b)~unknown artist\rq s {\em Saint Peter}, Convent of Saint Mary, Zadar (12th century), c)~Girolamo Dai Libri\rq s {\em Saint Peter with keys} (1533), d)~Perugino\rq s {\em Delivery of the keys to Saint Peter}, Vatican (1482), and e)~Peter Paul Rubens\rq\ {\em Saint Peter} (c.~1611).  In works such as these, Saint Peter is recognized by the presence of his attribute, keys.}
\end{figure}

Figure~\ref{fig:StPeter} shows five paintings of Saint Peter from different art historical periods, all showing his attribute, keys.  All viewers familiar with Christian iconography recognize that these paintings depict the same figure, despite significant differences in size, style, lighting, pose, and scene, all because of the presence of the attribute, itself portrayed amidst numerous size and stylistic variations.  While the figure\rq s identity can also be indicated by his visage and beard, costume  (papal vestments and pallium), and in some cases setting, such visual properties can differ widely for a given saint and be quite similar for different saints.  In short, a symbol or attribute is often the most a reliable visual marker of many important saints in Christian art.

We note in passing that most mythical and religious art of nearly all religions and sects employs such symbols functioning as attributes do.  In Greek and Roman mythology, for instance, Cupid is symbolized by a small bow and arrow, Zeus by a lightning bolt, Athena by an owl, Hera by a peacock, Poseidon by a trident, Bacchus by a bunch of grapes, and so forth.\cite{Martin:16}  Most of these associations have roots in Ovid\rq s {\em Metamorphoses}.  Likewise in Hindu iconography Brahma is represented by a lotus, Saraswati by a veena (stringed musical instrument), Parvati by a lion, and so on.  Most of these symbols have roots in ancient Hindu texts such as the {\em Bhagavad Gita}.

We demonstrate an automatic method for identifying actors---as a step toward inferring the source story and hence meaning---in Western Biblical art.  We present background for this goal and approach in Sect.~\ref{sec:Goal}.  Then, in Sect.~\ref{sec:DataCollection} we discuss the collection of religious art images for analysis and in Sect.~\ref{sec:AttributeRecognizer} an overview of our method for classifying symbols and attributes within paintings.  Next, in Sect.~\ref{sec:AttributeDatabase} we mention briefly our database of associations between attributes and saints.  In Sect.~\ref{sec:TrainingTesting} we discuss the training and testing of our deep neural networks for recognizing and localizing attributes and segmenting figures.  We quantify our results in Sect.~\ref{sec:Results} and interpret the sources of inevitable errors.  In Sect.~\ref{sec:Conclusions} we summarize our conclusions and discuss next steps in extraction of meaning from artwork and related problems in high-level image analysis of art. 

To the best of our knowledge, the work presented here is the first explicit work on automatic recognition of significant actors in art by any means, specifically through the use of symbols or attributes.

\section{TECHNICAL GOAL AND APPROACH} \label{sec:Goal}

In broad overview, our goal is to develop an automated system of image analysis that can, given a Christian painting in the Western canon from the past six centuries, identify saints (or major \lq\lq actors\rq\rq ).  Our system relies on an {\sc attribute recognition module} for symbols and attributes, a {\sc semantic segmentation module}, to localize candidate figures (actors), and an {\sc attribute saint association database}, a database of attributes (signifiers) and association saints (signifieds).  

Our approach includes the following steps:

\begin{itemize}
\item train a deep neural network {\sc attribute recognition module} to recognize and localize attributes (religious symbols associated with saints) from a small set of common such attributes, as listed in Table~\ref{tbl:SaintsAttributes}.
\end{itemize}

\noindent Given a painting to be analyzed:

\begin{itemize}
\item use a {\sc semantic segmentation module} (deep neural network) to segment the painting image, specifically to identify human actors
\item use the trained {\sc attribute recognition module} (deep neural network) to identify attributes and their locations within the artwork
\item find the nearest segmented human figure to each identified attribute
\item use the {\sc attribute saint association database} to identify the actor based on its closest semiotic attribute
\end{itemize}

\section{IMAGE COLLECTION AND DATA PREPARATION} \label{sec:DataCollection}

We scraped images of fine art Christian paintings from the 13th through 19th centuries in major museums, hand picked using search terms such as saints\rq\ names, the attributes, \lq\lq painting,\rq\rq\ \lq\lq mural,\rq\rq\ \lq\lq fresco,\rq\rq\ \lq\lq altarpiece,\rq\rq\ \lq\lq Medieval\rq\rq\ \lq\lq Renaissance,\rq\rq\ \lq\lq Baroque,\rq\rq\ and so forth.  We confirmed by eye that indeed each work depicted at least one saint and at least one attribute.  We hand classified attributes but not actors (saints).  We scaled all images to the same total number of pixels, to match the number of neurons in the input and output layers of our network (see Sects.~\ref{sec:AttributeRecognizer}--\ref{sec:TrainingTesting}).

\section{ATTRIBUTE RECOGNITION MODULE} \label{sec:AttributeRecognizer}

We trained a well-studied the convolutional architecture,\cite{Heetal:17} which stacks a feature pyramid network on top of a deep residual network \cite{Heetal:16} for region of interest extraction and binary pixel classification and segmentation.  We used the pre-trained weights trained by on the open-source common objects in context dataset.\cite{Heetal:17,Linetal:14}  For instance, we trained the network with images of  the \lq\lq keys to heaven\rq\rq\ depicted in the art in question.  

\section{ATTRIBUTE SAINT ASSOCIATION DATABASE} \label{sec:AttributeDatabase}

We created our  attribute database by hand, using scholarly sources of Christian iconography.\cite{Ferguson:90}   Table~\ref{tbl:SaintsAttributes} shows the several saints and attributes used in our study as the {\sc attribute saint association database}.  This list is of course not complete, either in the number of saints or the full complement of attributes for some saints.

\begin{table}[h]
\begin{center}
\begin{tabular}{|r|l|} \hline 
Saint & Attribute  \\ \hline
Christ & cross  \\
Matthew & angel  \\
Mark & winged lion  \\
Luke & bull  \\
Simon & boat \\
Thomas & ax \\
Catherine & wheel \\
Daniel & lion \\
George & dragon \\
John & eagle  \\ \hline
\end{tabular}
\end{center}
\caption{Several saints frequently depicted in Western Christian art and their leading attributes, stored as associations in our {\sc attribute association database}.  Several of the most-important saints have multiple attributes, for instance, Saint Peter\rq s attributes include keys of heaven, fish, boat, rooster, pallium and papal vestments, as discussed in Sect.~\ref{sec:Conclusions}.}
\label{tbl:SaintsAttributes}
\end{table} 

\begin{figure}[h] 
\begin{center}
\begin{tabular}{cc}
\includegraphics[width=.47\textwidth]{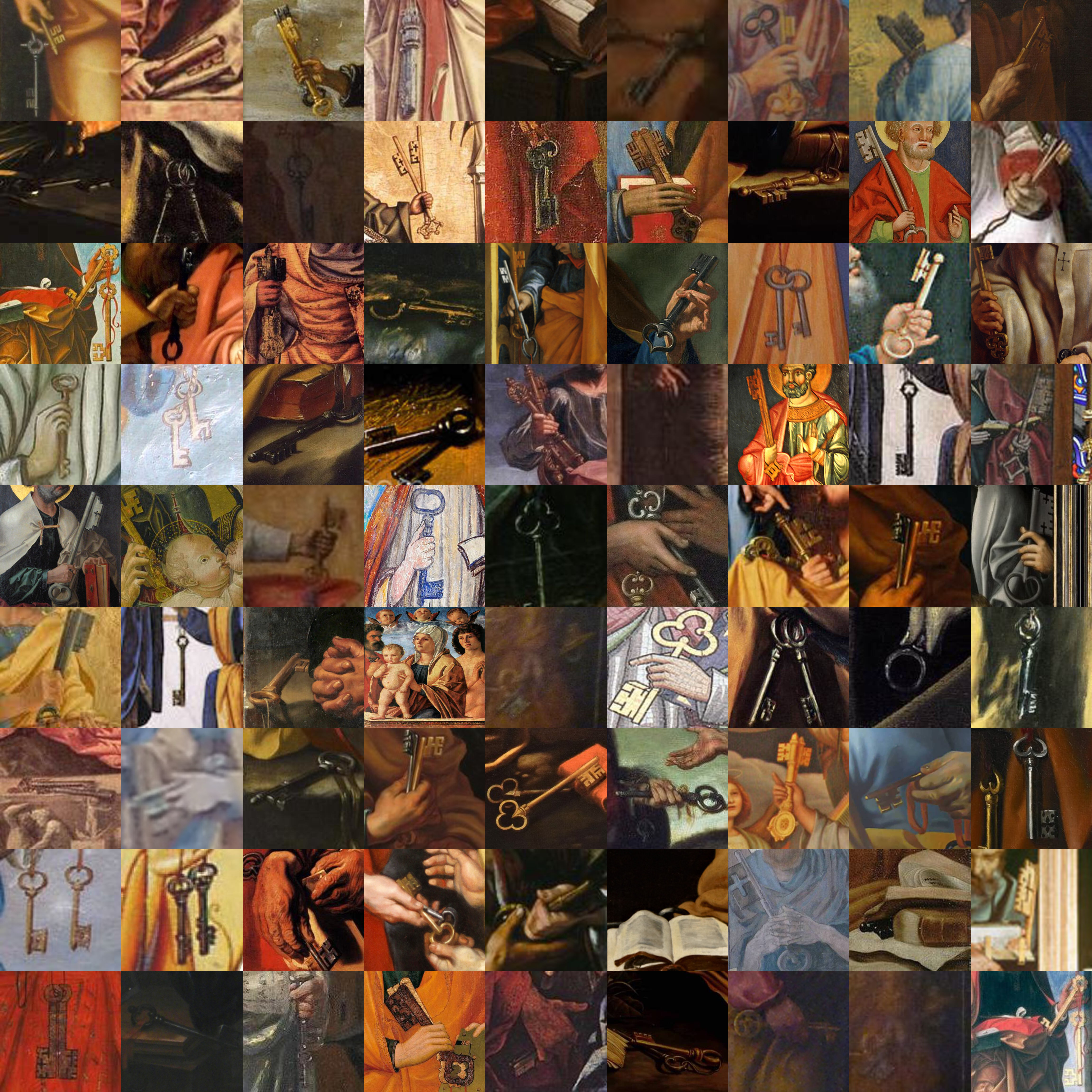} &
\includegraphics[width=.47\textwidth]{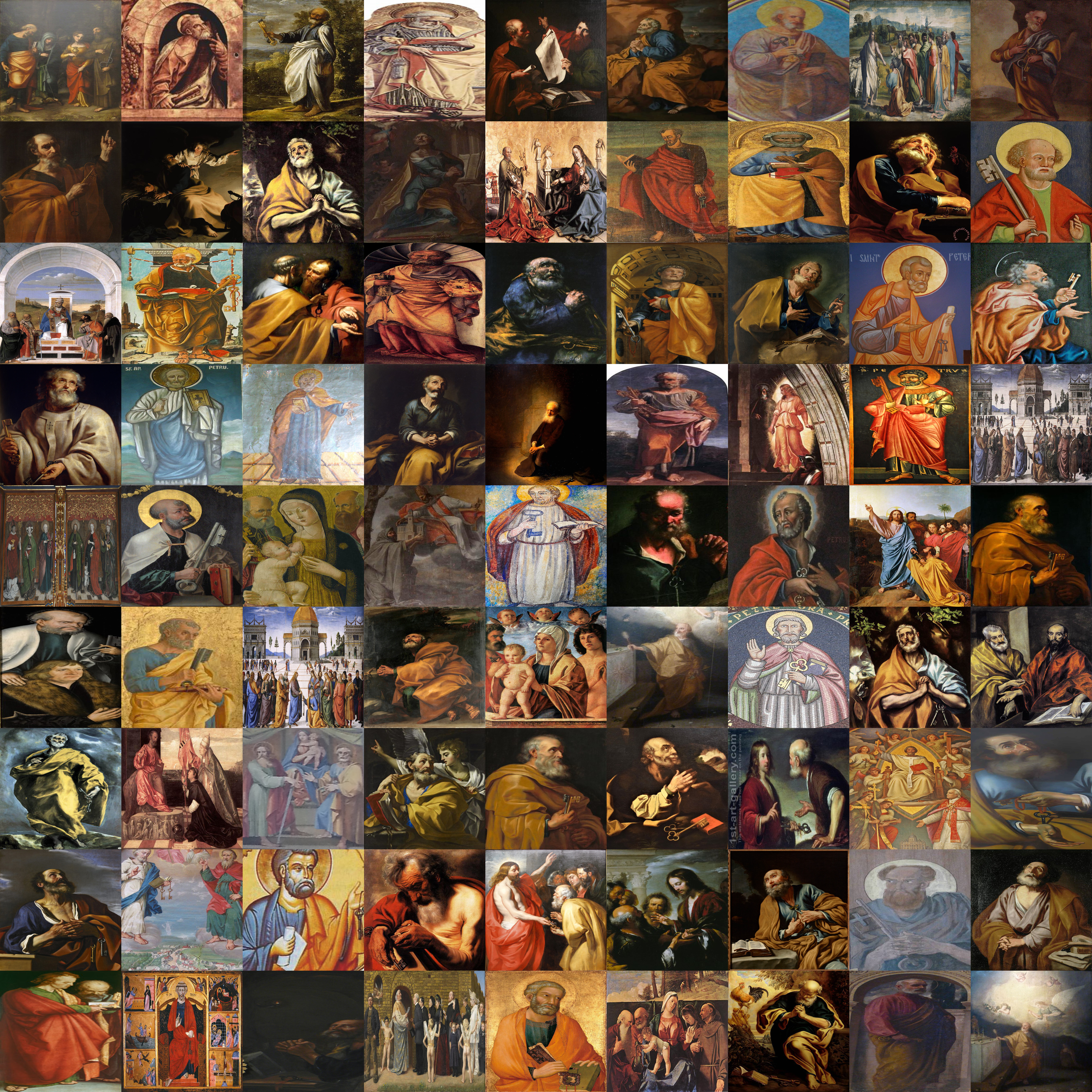} 
\end{tabular}
\end{center}
\caption{\label{fig:KeysSaintPeter}({\sc L})~Images of the attribute of keys culled from Western artwork from the 13th through 18th centuries, used for learning of Saint Peter\rq s attribute.  ({\sc R})~Images of Saint Peter, most of which include his attribute.}
\end{figure}

\section{TRAINING AND TESTING} \label{sec:TrainingTesting}

We used well-established protocols of leave-one-out training to ensure that any test image was not used during training.\cite{DudaHartStork:01}  Our ground truth was provided by independent references in art history and Christian iconography and the titles of the artworks analyzed.\cite{Ferguson:90,ApostolosCappadona:20}  Test paintings depicting saints from Table~\ref{tbl:SaintsAttributes} were scraped from online art databases based on keywords and contextual phrases, as described in Sect.~\ref{sec:DataCollection}.  Figure~\ref{fig:KeysSaintPeter} shows examples of keys as well as Saint Peter holding keys.  Notice the great variety in sizes, styles, colors, orientations of such keys.  Of course the training and test sets were disjunct:  we never tested our attribute recognizer on a painting used for its training.\cite{DudaHartStork:01}  We downsampled each image so it matched the pixel (neuron) numbers in the semantic segmentation network.  A simple routine computed the pixel location of the center of mass of each figure.

Next, a very simple routine identified the location of the figure nearest to each attribute, according to the figure\rq s visual center of mass.  The final step was to assign the saint in the {\sc attribute saint association database} in Table~\ref{tbl:SaintsAttributes} to that closest figure.  Testing the overall system consisted of presenting a novel Christian painting, classifying each significant actor,  and comparing the result to the identification provided by a human expert.  In this way we find the percentage of true and false positives, as well as true and false negatives.

\section{RESULTS AND INTERPRETATIONS} \label{sec:Results}

\begin{figure}[h] 
\begin{center}
\begin{tabular}{cc}
\includegraphics[width=.45\textwidth]{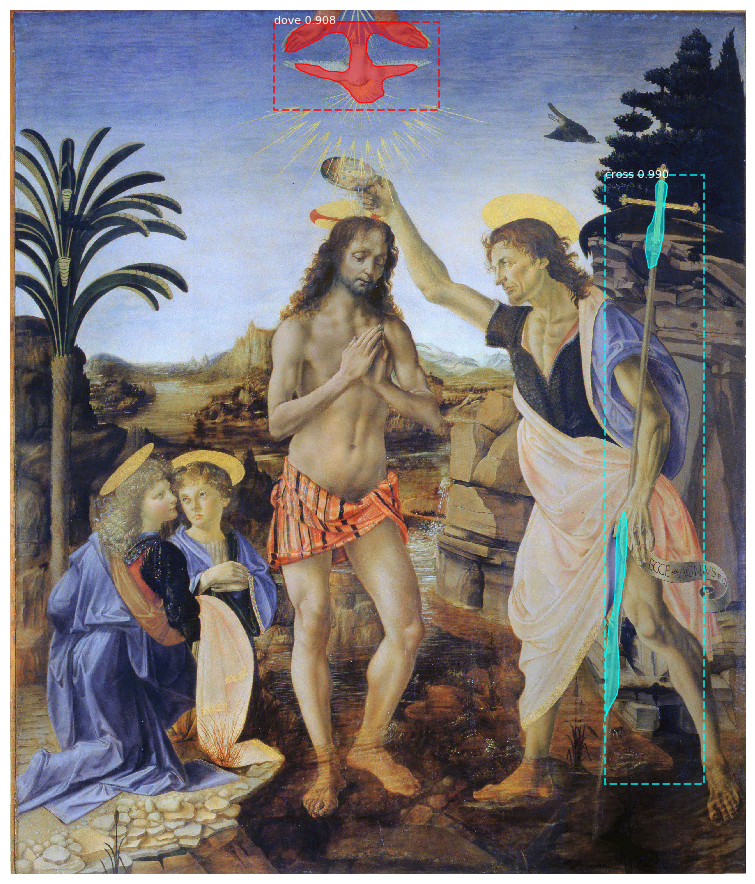} &
\includegraphics[width=.45\textwidth]{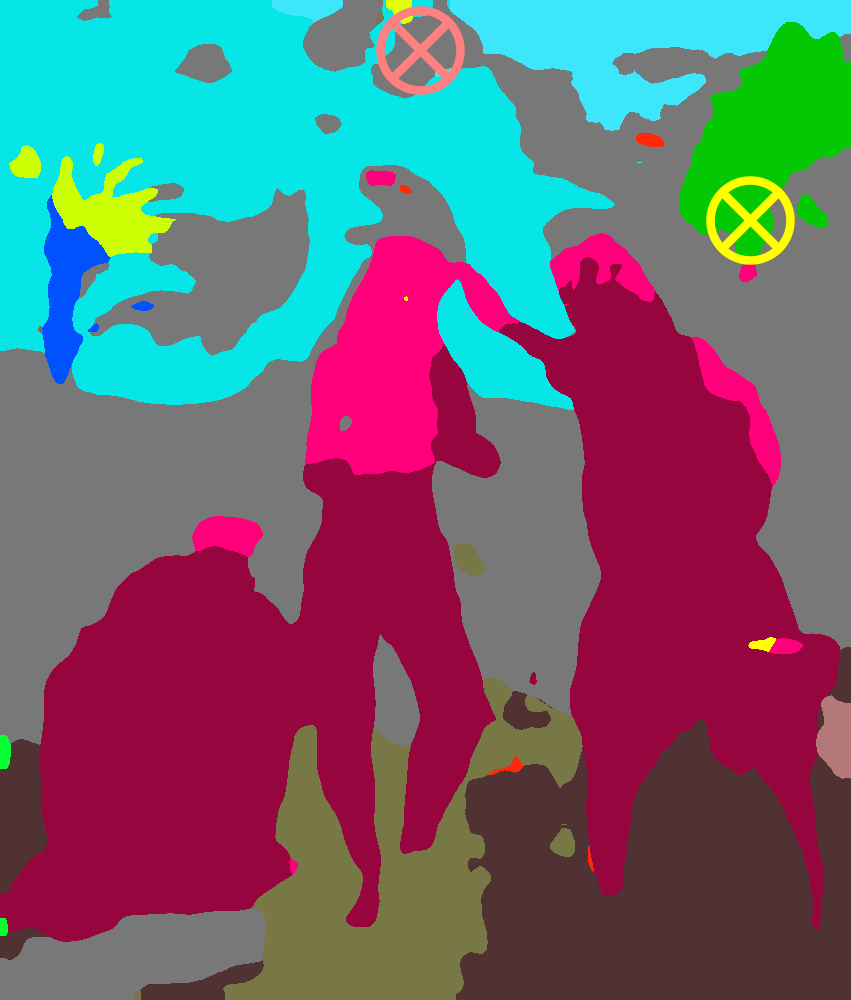} 
\end{tabular}
\end{center}
\caption{\label{fig:VerrocchioInterpretation}({\sc L})~Andrea Verrocchio\rq s {\em Baptism of Christ} ($177 \times 151\ cm$) oil on wood (1475), Uffizi Gallery, Florence, and bounding boxes of saints\rq\ attributes found automatically.  Here, the dove and cross are each found with a confidence of $99\%$.  ({\sc R})~The semantic segmentation where the pink and crimson indicate {\sc animal} and {\sc person}, which jointly define the figures.  The proximity of the attributes to the figures properly identify the central figure as Christ and the figure at the right as Saint John the Baptist, the correct reading of this work.  Notice the presence of unphysical items, such as halos and golden rays emanating from above.}
\end{figure}

Figure~\ref{fig:VerrocchioInterpretation} shows the analysis by our trained system on Andrea Verrocchio\rq s {\em Baptism of Christ}.  The left panel shows the output of the {\sc attribute recognition module}, which identifies which attributes are most confidently detected---here of {\sc dove} and {\sc cross}---each with an extremely high confidence of $99\%$.  The {\sc attribute recognizer module} was applied to each test image to yield the locations and statistical confidences associated that each candidate attribute was present at a given location in the test painting.  Only the one or two attributes whose presence was computed to be above a high threshold were retained, which thus yielded a spatial location for that attribute.  The network also localizes each such attribute, as indicated by the bounding boxes. The right panel shows the output of the {\sc semantic segmentation module}.  Note especially the segmentation regions for {\sc person} and {\sc animal}.  Our algorithm merges such regions into a single region, which was then a candidate actor.  As is evident from that panel, the semantic segmentation is quite accurate throughout the dataset.  We can expect that segmentation will be even more accurate and robust if the network is trained on a larger number of paintings representative of the works in the test set.

\begin{figure}[h] 
\begin{center}
\begin{tabular}{cccc}
\includegraphics[height=3.7cm]{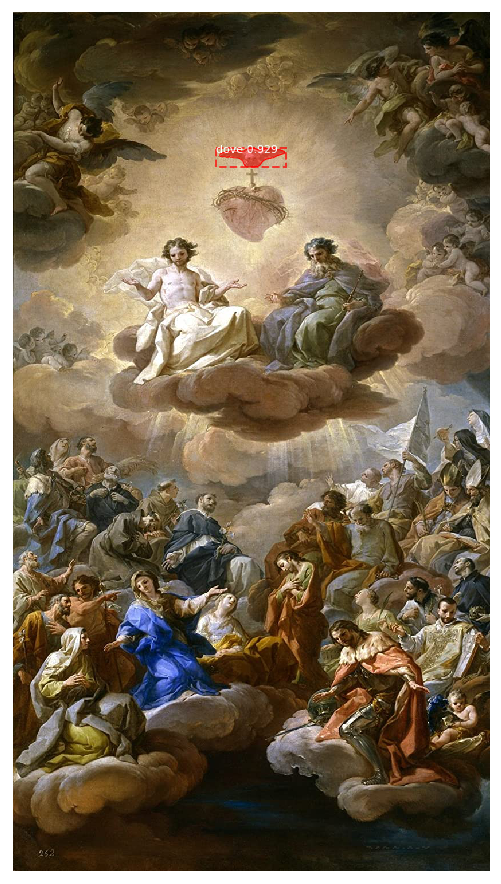} &
\includegraphics[height=3.7cm]{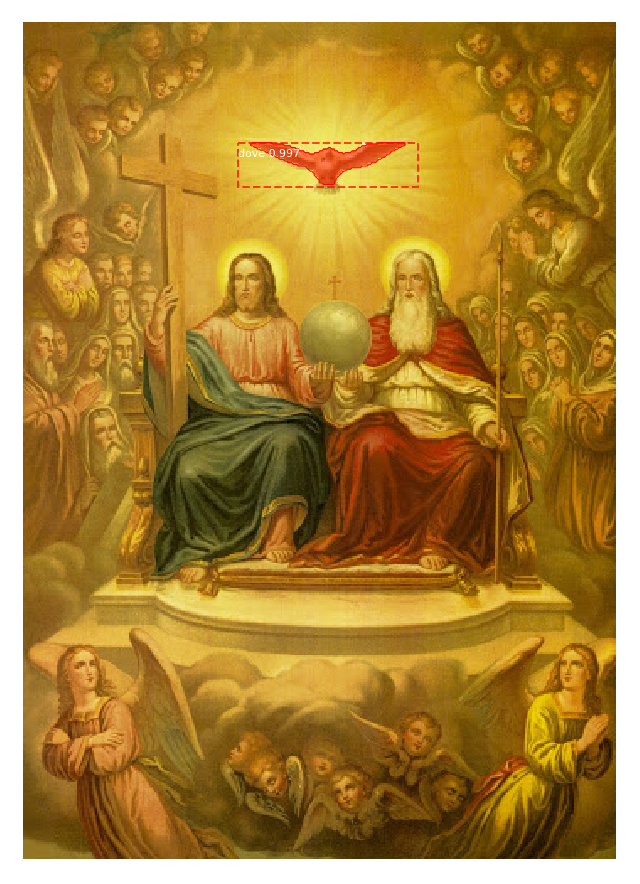} &
\includegraphics[height=3.7cm]{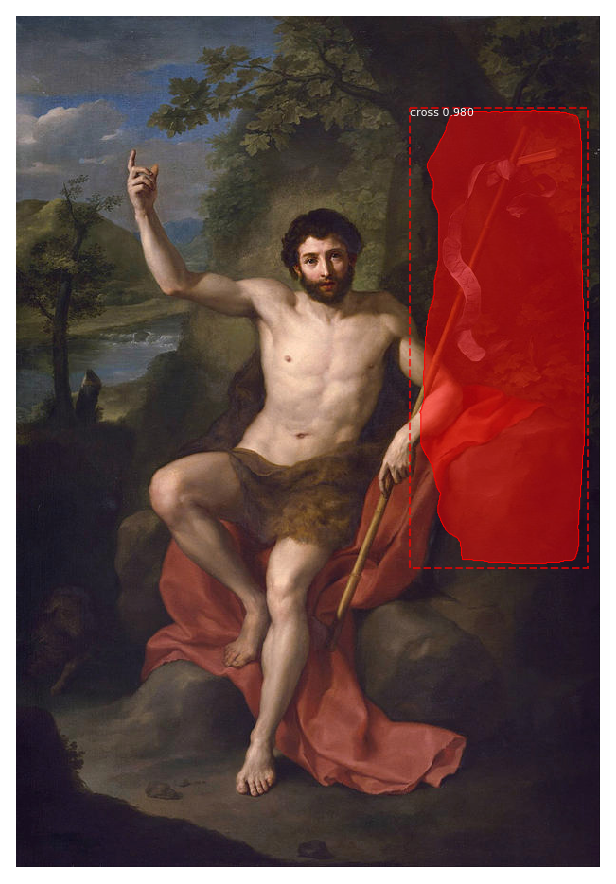} &
\includegraphics[height=3.7cm]{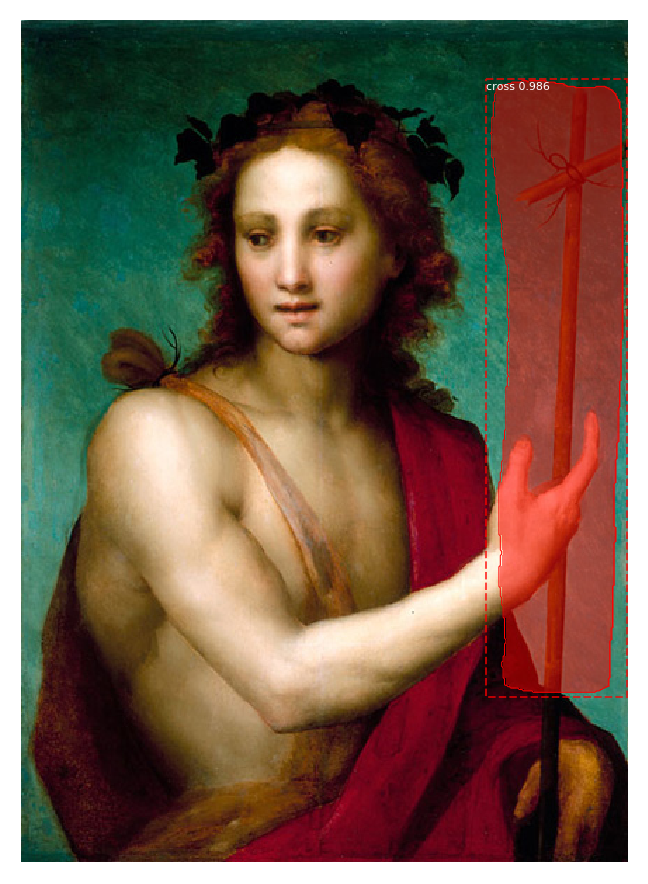} \\
a) & b) & c)~ & d) \\
\includegraphics[height=3.7cm]{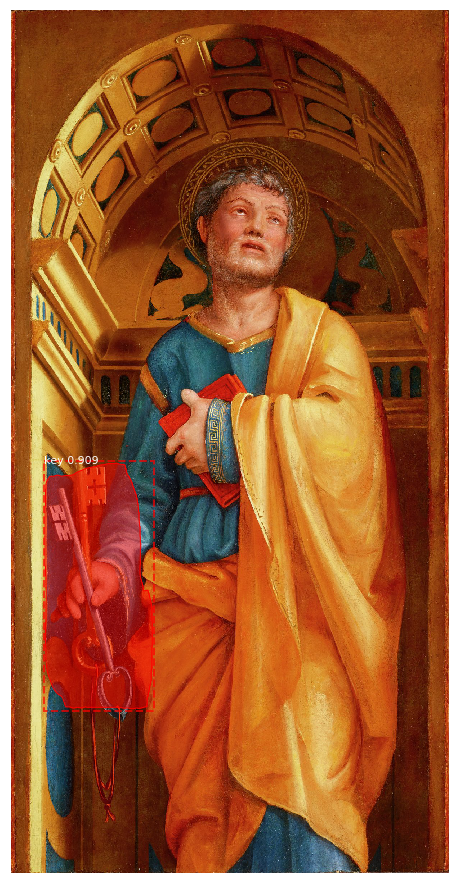} &
\includegraphics[height=3.7cm]{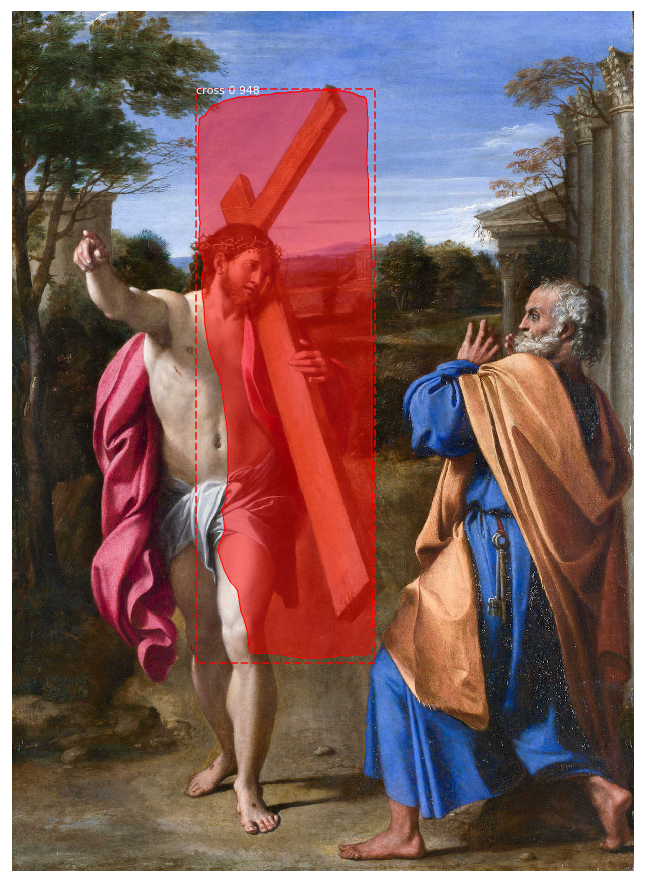} &
\includegraphics[height=3.7cm]{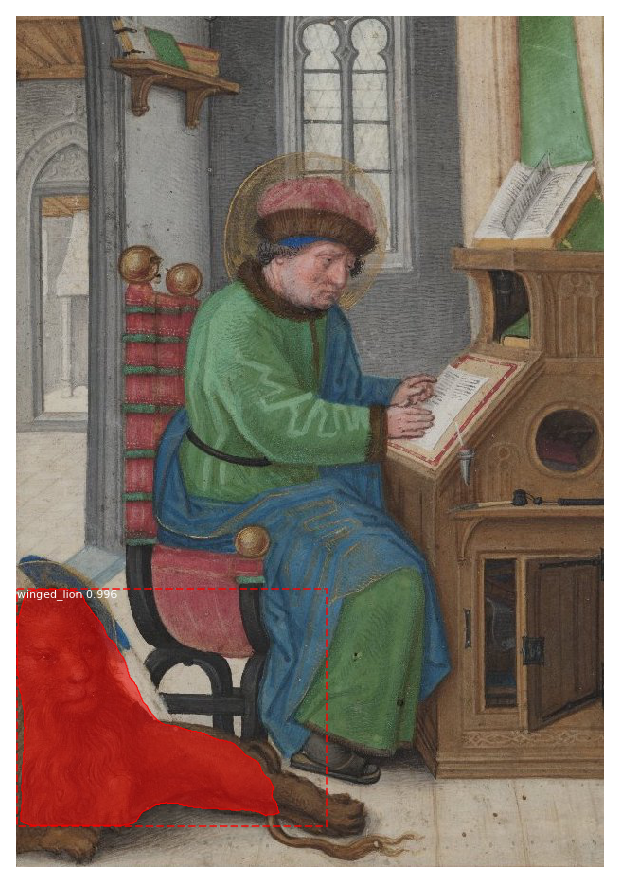} &
\includegraphics[height=3.7cm]{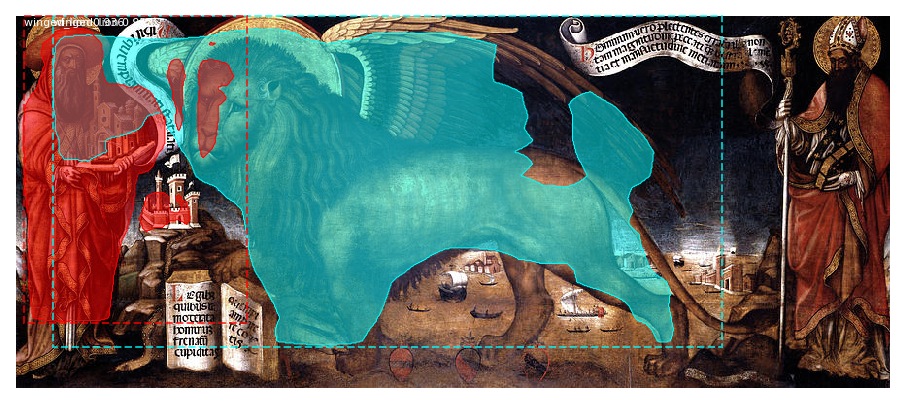} \\
e) & f) & g) & h)
\end{tabular}
\end{center}
\caption{\label{fig:Examples}The output of the {\sc attribute recognition module}, here accurately finding attributes in the following paintings:  a)~Corrado Giaquinto\rq s {\em The holy trinity} (c.~1755), b)~unknown artist {\em Holy trinity} (unknown), c)~Anton Raphael Mengs\rq\ {\em St.~John the Baptist preaching the wilderness} (1760s), d)~Andrea del Sarto\rq s {\em Saint John the Baptist} (c.~1517), e)~Bernardino Zenale\rq s {\em Saint Peter the apostle} (c.~1510--12), f)~Annibale Carracci\rq s {\em Christ appearing to Saint Peter on the Appian Way} (1601--02), g)~Simon Bening\rq s {\em St.~Mark writing} (1521), and h)~Donato Veneziano\rq s {\em The lion \lq\lq andante\rq\rq\ of St.~Mark} (1459).}
\end{figure}

Figure~\ref{fig:Examples} shows eight paintings in our dataset and the output of the {\sc attribute classifier module}, specifically two paintings showing the dove, four showing the crucifixion cross, and two showing a lion.  Note particularly that the crucifixion cross can be associated with Christ of Saint John (and yet other saints).  We then automatically computed the spatial distance between the center of mass of each segmented candidate actor and the location of the confidently identified semiotic attributes.  The identity of the actor was assigned that associated with its closest such attribute according to the {\sc attribute saint database} in Table~\ref{tbl:SaintsAttributes}.  For example, in {\em Baptism of Christ}, above, the attributes {\sc Dove} and {\sc cross} were associated with Christ and Saint John the Baptist, respectively---a proper \lq\lq reading\rq\rq\ of this painting.

The system performance is expressed as precision and recall, where {\em precision} represents the proportion of the identifications are actually correct and {\em recall} represents the proportion actually positives that were correctly identified by the system.  Formally, these statistics are defined by:

\begin{equation} \label{eq:PrecisionRecallDef}
{\rm precision} = \frac{TP}{TP + FP}\quad\quad {\rm and}\quad\quad  {\rm recall} = \frac{TP}{TP + FN} ,
\end{equation}

\noindent where $TP$ is represents true positives, $FP$ false positives, and $FN$ false negatives.  Table~\ref{tbl:PrecisionRecall} shows the precision and recall for our system based on the results in Fig.~\ref{fig:Examples}.

\begin{table}
\begin{center}
\begin{tabular}{|r|cccc|} \hline
 & God & Mark & John & Peter \\ \hline
 Precision & $100\%$ & $67\%$ & $100\%$ & $100\%$ \\
 Recall  & $100\%$ & $100\%$ & $75\%$ & $50\%$ \\ \hline
\end{tabular}
\end{center}
\caption{\label{tbl:PrecisionRecall}The precision and recall of our system for four saints.}
\end{table}

\section{CONCLUSIONS AND FUTURE DIRECTIONS} \label{sec:Conclusions}

Our overarching program is to develop computer-based methods for the interpretation of fine art, in particular to extract rudimentary meanings.  Art images differ in many ways from natural photographs, which dominate research into semantic image analysis research.  Art explores an extraordinarily larger range of styles than do photographs and unlike most photographs art is frequently crafted by an artist (\lq\lq author\rq\rq ) seeking to express a story, concept, moral, or meaning.  As such, it is important that automated system identify not merely the picture\rq s components and their relations (as is common in traditional semantic image analysis), but also what the image {\em means}.

We have demonstrated that one early step in the interpretation of story or meaning in paintings in the Western canon can be computed automatically.  Specifically, our system can reliably identify the actors (here, saints) in Christian art by means of their symbols or attributes.  The system relies on special-purpose {\sc attribute classifier module}, existing {\sc semantic segmentation module} networks, and a curated associative {\sc attribute saint association database}.  

There are a number of complexities and challenges in expanding this work to larger corpora, even if restricted to religious art in the Western canon.  First, the total number of saints with attributes depicted in artwork is in the hundreds.  Some of these associated attributes are complex and difficult to recognize, such as silk gloves---the attribute of Saint Louis of Toulouse.  The number and variety of attributes or symbols is likely equally large.  The computational task becomes even more challenging because multiple saints share the same attributes.  For those cases, the attribute taken alone is not sufficient for accurate actor identification.  A more probabilistic inference may be necessary, one based on some global criterion of all the actors and their simultaneous presence in candidate stories or episodes.

Recall that our {\sc attribute saint association database} was hand curated for this preliminary proof-of-concept work.  Future work based on natural language analysis of source texts, such as titles of artworks, Biblical source stories, and art historical analyses may be sufficient to learn the associations between saints and attributes.  A great benefit of this approach would be to learn concurrence probabilities, which would likely improve the accuracy of the overall system.  Of course all results are probabilistic, so this level of analysis may yield two or three candidate saints, each with a confidence score.  In such cases, the presence of other actors in the scene---each with its own confidence score---can likely lead to an improved recognition of each {\em overall}.  Although existing segmentation methods were adequate for the work presented here, we can expect improved performance on challenging large datasets if transfer learning is used with representative art images themselves.


\section*{ACKNOWLEDGEMENTS}

The first author would like to thank the Getty Research Center for access to its Research Library, where some of the above research was conducted.  We also thank Grace Bourached and Rosie Bourached for assistance in data preparation.

\bibliography{Art}
\bibliographystyle{spiebib}
 
\end{document}